\documentclass[Numbered,sn-basic]{sn-jnl}

\usepackage{graphicx}%
\usepackage{multirow}%
\usepackage{amsmath,amssymb,amsfonts}%
\usepackage{amsthm}%
\usepackage{mathrsfs}%
\usepackage[title]{appendix}%
\usepackage{textcomp}%
\usepackage{manyfoot}%
\usepackage{booktabs}%
\usepackage{algorithm}%
\usepackage{algorithmicx}%
\usepackage{algpseudocode}%
\usepackage{listings}%
\usepackage{soul}%
\usepackage{xcolor}%
\sethlcolor{yellow}
\soulregister{\cite}{7}
\soulregister{\citep}{7}
\soulregister{\citet}{7}
\soulregister{\ref}{7}
\soulregister{\pageref}{7}
\soulregister{\eqref}{7}
\usepackage{amsmath}
\usepackage{appendix}
\usepackage{bm}
\usepackage{tabularx}

\theoremstyle{thmstyleone}%

\theoremstyle{thmstyletwo}%

\theoremstyle{thmstylethree}%

\begin{document}

\title[Article Title]{Optimal Pose Guidance for Stereo Calibration in 3D Deformation Measurement}

\author[1,2]{\fnm{Dongcai} \sur{Tan}}

\author[1,2]{\fnm{Shunkun} \sur{ Liang}}

\author[1,2]{\fnm{Bin} \sur{Li}}

\author*[1,2]{\fnm{Banglei} \sur{Guan}}\email{banglei.guan@hotmail.com}

\author[1,2]{\fnm{Ang} \sur{Su}}

\author[3]{\fnm{Yuan} \sur{Lin}}

\author[1,4]{\fnm{Dapeng} \sur{Zhang}}

\author[3]{\fnm{Minggang} \sur{Wan}}

\author[1,2]{\fnm{Zibin} \sur{ Liu}}

\author[3]{\fnm{Chenglong} \sur{Wang}}

\author[3]{\fnm{Jiajian} \sur{Zhu}}

\author[1,2]{\fnm{Zhang} \sur{Li}}

\author[1,2]{\fnm{Yang} \sur{Shang}}

\author[1,2]{\fnm{Qifeng} \sur{Yu}}

\affil[1]{\orgdiv{College of Aerospace Science and Engineering}, \orgname{National University of Defense Technology}, \orgaddress{\city{Changsha}, \postcode{410073}, \state{Hunan}, \country{China}}}

\affil[2]{\orgdiv{Hunan Provincial Key Laboratory of Image Measurement and Vision Navigation}, \orgaddress{\city{Changsha}, \postcode{410073}, \state{Hunan}, \country{China}}}

\affil[3]{ \orgdiv{Hypersonic Technology Laboratory}, \orgname{National University of Defense Technology}, \orgaddress{\city{Changsha}, \postcode{410073}, \state{Hunan}, \country{China}}}

\affil[4]{\orgdiv{Hunan Key Laboratory of Intelligent Planning and Simulation for Aerospace Missions}, \orgaddress{\city{Changsha}, \postcode{410073}, \state{Hunan}, \country{China}}}

\abstract{\textbf{Background} Stereo optical measurement techniques, such as digital image correlation (DIC), are widely used in 3D deformation measurement as non-contact, full-field measurement methods, in which stereo calibration is a crucial step. However, current stereo calibration methods lack intuitive optimal pose guidance, leading to inefficiency and suboptimal accuracy in deformation measurements. 

\textbf{Objective} The aim of this study is to develop an interactive calibration framework that automatically generates the next optimal pose, enabling high-accuracy stereo calibration for 3D deformation measurement.

\textbf{Methods} We propose a pose optimization method that introduces joint optimization of relative and absolute extrinsic parameters, with the minimization of the covariance matrix trace adopted as the loss function to solve for the next optimal pose. Integrated with this method is a user-friendly graphical interface, which guides even non-expert users to capture qualified calibration images.

\textbf{Results} Our proposed method demonstrates superior efficiency (requiring fewer images) and accuracy (demonstrating lower measurement errors) compared to random pose, while maintaining robustness across varying FOVs. In the thermal deformation measurement tests (500–700°C) on an S-shaped specimen,  the results exhibit high agreement with finite element analysis (FEA) simulations in both deformation magnitude and evolutionary trends.

\textbf{Conclusions} We present a pose guidance method for high-precision stereo calibration in 3D deformation measurement. The simulation experiments, real-world experiments, and thermal deformation measurement applications all demonstrate the significant application potential of our proposed method in the field of 3D deformation measurement.}

\keywords {Stereo calibration, Optimal pose guidance, 3D deformation measurement, Digital image correlation}

\maketitle

\section{Introduction}\label{sec1}

Digital image correlation \cite{bib1, bib2, bib3}, and other optical measurement techniques \cite{bib4, bib5, bib6, bib7, bib8, lei} have emerged as standard solutions for 3D reconstruction, deformation analysis, and pose estimation due to their non-contact, full-field measurement, and subpixel accuracy. These techniques are widely adopted in industrial inspection \cite{bib9, bib10}, biomechanics \cite{bib11, bib12}, and extreme mechanics \cite{bib13, bib14}. Camera calibration is a key step for optical measurement techniques, which is crucial for achieving accurate image measurement. Essentially, camera calibration is the process of establishing a mapping model between the 3D control points and corresponding image points \cite{bib15, bib16, bib17, bib18}. Both MATLAB and OpenCV have integrated calibration toolboxes \cite{bib19, bib20}, which require users to manually select the poses of the calibration target and then optimize the camera parameters from a batch of calibration images by estimating the maximum likelihood. However, achieving robust and high-precision camera calibration requires professional calibration experience to precisely control the pose of the calibration target or the camera. It is well known that to accurately estimate the distortion coefficients and camera intrinsic parameters, the calibration target should cover as large a field of view as possible. And to avoid the degraded pose of the camera, Reimar \cite{bib21} and Zhang \cite{bib22} recommend varying the relative orientation between the camera and the calibration target as much as possible when acquiring calibration images. However, these experiences and studies do not intuitively guide inexperienced users in acquiring calibration images that can yield high-precision calibration results.

Some scholars have investigated methods for guiding users through the camera calibration process. Tan et al. \cite{bib23} project virtual patterns onto the screen, thus avoiding the operator from moving the camera or marker, but they did not study how to control the pose relationship between the virtual pattern and the camera for efficient calibration. Tsoy et al. \cite{bib25} proposed a virtual camera calibration evaluation pipeline, while the study automates pose sampling within the viewing frustum, it lacks explicit optimization of target-camera pose relationships to minimize reprojection errors, akin to the limitation noted by \cite{bib24}  in virtual pattern calibration. Some studies have employed the minimization of the uncertainty of the internal parameter estimation \cite{bib26}, geometric reprojection error \cite{bib27, bib28}, and the uncertainty of corner points \cite{bib28} as loss functions to search for the next optimal pose, thus guiding the users. However, the methods mentioned above are all applicable only to monocular calibration. Compared to monocular systems, stereo vision systems exhibit superior depth sensitivity and measurement accuracy due to their inherent epipolar geometry and binocular disparity constraints. 

In response to these problems, we propose a pose optimization algorithm and a system that guides users to move the calibration target to the optimal pose for high-precision stereo calibration. On one hand, our approach introduces a coupled optimization term for the relative and absolute extrinsic parameters, which enables joint geometric parameter estimation that surpasses the capabilities of monocular methods lacking stereometric constraints. On the other hand, while DIC technology is mature, expertise in optimizing calibration target poses for high-accuracy calibration remains limited. Our method empowers non-specialists to achieve high-precision calibration efficiently, reducing DIC’s application barrier with tangible practical value. Our contributions to this paper can be summarized as follows:
\begin{itemize}
  \item   We propose a method that solves for the next optimal pose by introducing a joint optimization for relative and absolute extrinsic parameters and adopting the minimization of the trace of the covariance matrix as the loss function.
  \item   We present a user-friendly system for generating optimal poses tailored to high-precision stereo calibration, which guides users even without prior expertise to capture qualified calibration images via an intuitive graphical interface. 
  \item  Simulation experiments, real-world comparative experiments, and validation in thermal deformation application confirm that the proposed method is practically viable for stereo calibration, characterized by high precision and efficiency.
\end{itemize}

The remainder of the paper is organized as follows. Section \ref{sec2} introduces the stereo imaging model. Section \ref{sec3} describes the optimal pose guidance process and mathematical details of the proposed method. The comparative experiments and thermal deformation experiment are reported in Section \ref{sec4}. And section \ref{sec5} draws the conclusions.

\section{Stereo Imaging Model}\label{sec2}

\begin{figure}[t]
\centering
\includegraphics[width=0.6\textwidth]{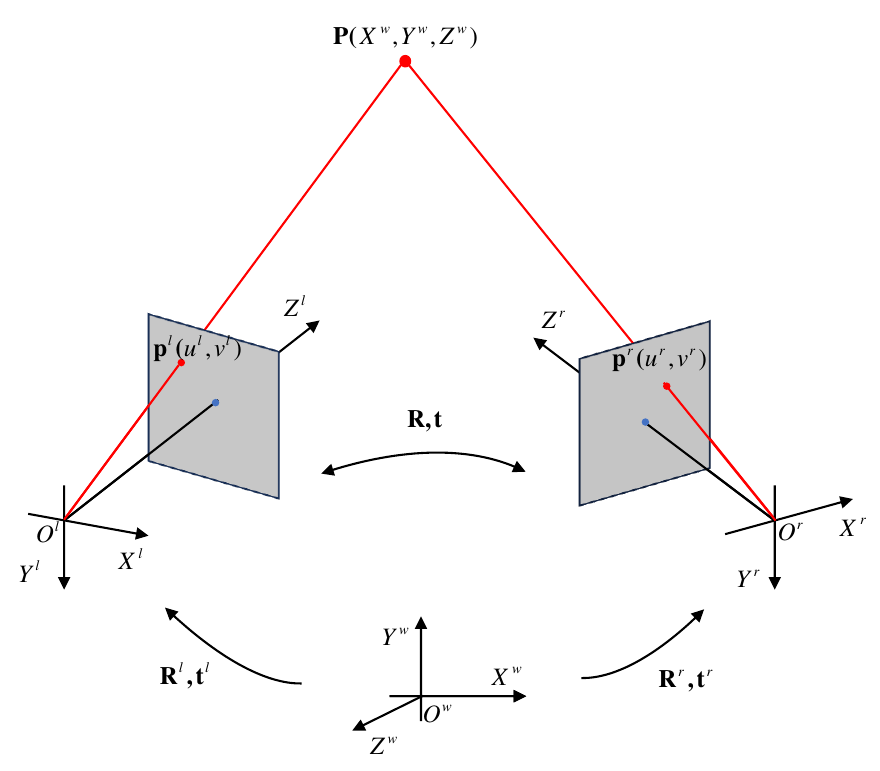}
\caption{Stereo imaging model}\label{fig1}
\end{figure}

As shown in Fig. \ref{fig1}, the stereo imaging model is typically composed of two cameras with a common field of view (FOV), which maps a 3D point in the real world to the left and right camera planes. A 3D control point $\mathbf{P}({X^w},{Y^w},{Z^w})$ in the world coordinate system can be represented in homogeneous coordinates as $\mathbf{P} = {[{X^w},{Y^w},{Z^w},1]^T}$, and it can be projected onto the left and right camera planes by the homogeneous coordinates, denoted as ${\mathbf{p}^l} = {[{u^l},{v^l},1]^T}$ and ${\mathbf{p}^r} = {[{u^r},{v^r},1]}^T$, respectively. The projection calculation process is as follows:
\begin{equation}
{s^l}\left[ {\begin{array}{*{20}{c}}
{{u^l}}\\
{{v^l}}\\
1
\end{array}} \right] = {\mathbf{K}^l}\left[ {{\mathbf{R}^l}|{\mathbf{t}^l}} \right]\left[ {\begin{array}{*{20}{c}}
{{X^w}}\\
{{Y^w}}\\
{{Z^w}}\\
1
\end{array}} \right]
\label{equ1}
\end{equation}

\begin{equation}
{s^r}\left[ {\begin{array}{*{20}{c}}
{{u^r}}\\
{{v^r}}\\
1
\end{array}} \right] = {\mathbf{K}^r}\left[ {{\mathbf{R}^r}|{\mathbf{t}^r}} \right]\left[ {\begin{array}{*{20}{c}}
{{X^w}}\\
{{Y^w}}\\
{{Z^w}}\\
1
\end{array}} \right]
\label{equ2}
\end{equation}

\noindent where $s^l$ and $s^r$ denote the depth scaling factors, which help to convert pixel coordinates in an image into actual physical units. $\mathbf{K}^l$ and $\mathbf{K}^r$ are the intrinsic parameter matrices of the cameras, which reflect the geometric structure and optical characteristics of the camera, and are used to project 3D camera coordinates onto the 2D pixel coordinates. The rotation matrices $\mathbf{R}^l$ and $\mathbf{R}^r$ and the translation vectors $\mathbf{t}^l$ and $\mathbf{t}^r$ are referred to as the absolute extrinsic parameters, which denote the rigid transformation from the world coordinate system to the camera coordinate system. 

It is important to note that the left and right cameras are not entirely independent. Once the stereo measurement system is set up, the relative extrinsic parameters between the left and right cameras are fixed. The absolute extrinsic parameters of the right camera can be expressed through the absolute extrinsic parameters of the left camera and the relative extrinsic parameters:
\begin{equation}
\left\{
\begin{aligned}
&\mathbf{R}^r = \mathbf{R}\mathbf{R}^l \\
&\mathbf{t}^r = \mathbf{R}\mathbf{t}^l + \mathbf{t}
\end{aligned}
\right.
\label{equ3}
\end{equation}

\noindent where $\mathbf{R} \in SO(3)$ is the rotation matrix and $\mathbf{t} \in \mathbb{R}^3$ is the translation vector.

The stereo imaging model establishes the projective mapping between 3D control points and their corresponding undistorted 2D image points. Meanwhile, the distortion model \cite{bib29} refers to the geometric distortion of objects in images due to the non-ideal characteristics of the camera lens. Specifically, such distortions are primarily divided into two major categories: radial distortion \cite{bib30} and tangential distortion \cite{bib31}. The radial distortion coefficients $k_1$ and $k_2$ describe the degree of distortion along the radial direction from the image center. And the tangential distortion coefficients $p_1$ and $p_2$ describe the degree of distortion along the tangential direction. 

Let us represent the distortion coefficients by a $4 \times 1$ vector $\mathbf{d}$, that is, $\mathbf{d} = {[{k_1},{k_2},{p_1},{p_2}]^T}$. Typically,  reprojection error is used to quantify calibration accuracy, which is defined as the difference between the projected image points and the actual observed image points. Suppose that we use the left and right cameras to capture the corresponding $m$ pairs of images, and each pair of images has $n$ pairs of corresponding corner points. On this basis, the geometric reprojection error for stereo calibration can be expressed as:
\begin{equation}
\begin{aligned}
\mathop {\min }\limits_{\mathbf{X}_i} \sum_{i = 1}^m \sum_{j = 1}^n \Big[ 
    &\rho \left( \parallel \mathbf{p}_{ij}^l - \pi (\mathbf{K}^l,\mathbf{d}^l,\mathbf{R}_i^l,\mathbf{t}_i^l,\mathbf{P}_j) \parallel^2 \right) \\
    &+ \rho \left( \parallel \mathbf{p}_{ij}^r - \pi (\mathbf{K}^r,\mathbf{d}^r,\mathbf{R}_i^r,\mathbf{t}_i^r,\mathbf{P}_j) \parallel^2 \right) \Big]
\end{aligned}
\label{equ4}
\end{equation}

\noindent where $\mathbf{X}_i = \{ \mathbf{K}^l, \mathbf{d}^l,\mathbf{R}_i^l,\mathbf{t}_i^l, \mathbf{K}^r, \mathbf{d}^r, \mathbf{R}_i^r, \mathbf{t}_i^r \}$ denotes the set of parameters to be optimized, $\rho (\cdot)$ denotes the robust kernel function, and the projection equation is represented by $\pi (\cdot)$.

\section{Methodology}\label{sec3}

This paper proposes a pose optimization algorithm to compute the next optimal pose, which assists users in achieving high-precision and efficient stereo calibration. Section \ref{subsec31} briefly introduces the process for guiding stereo calibration, and Section \ref{subsec32} elaborates on the mathematical formulation of the proposed method.

\subsection{Stereo Calibration Guidance Process}\label{subsec31}

\begin{figure}[t]
\centering
\includegraphics[width=1\textwidth]{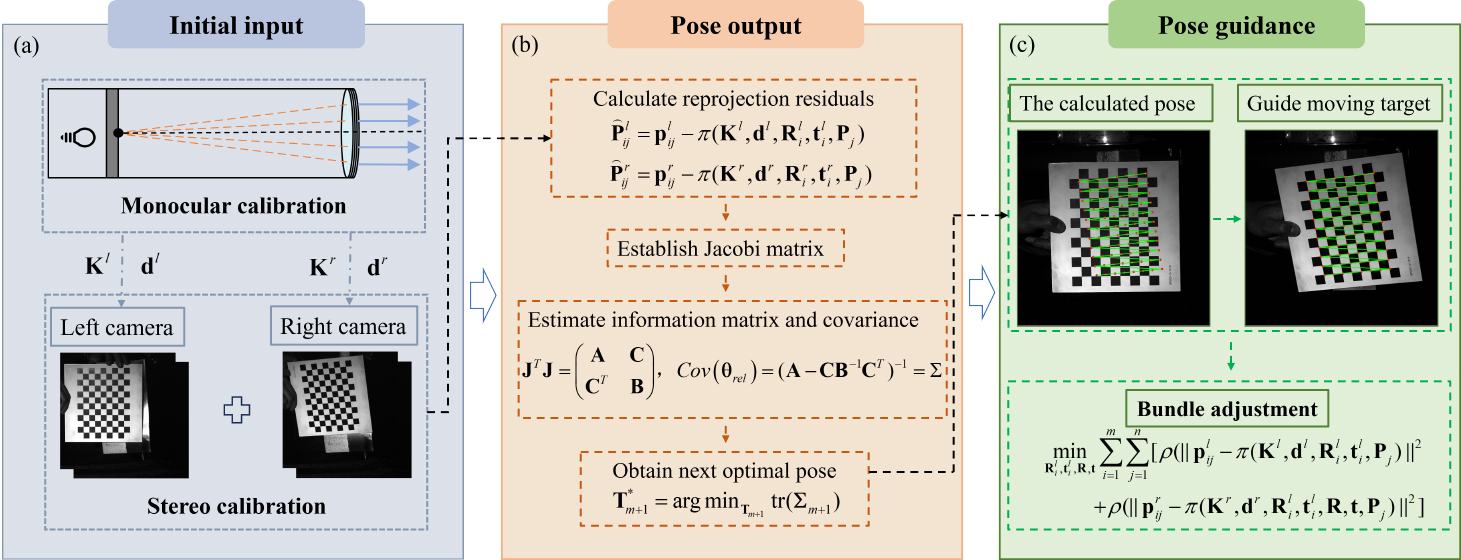}
\caption{Flowchart of the process of guiding the stereo calibration. \textbf{(a)} Initial input module, \textbf{(b)} pose output module, and \textbf{(c)} pose guidance module}
\label{fig2}
\end{figure}

The stereo calibration guidance process is illustrated in Fig. \ref{fig2}. The stereo calibration workflow is systematically divided into three modules: (1) Initial input module, where parameters of left and right cameras are determined while collecting calibration images to compute initial estimates of relative and absolute extrinsic parameters; (2) Pose output module, employing the established parameters to construct Jacobian matrices, and the optimal pose is determined by searching for the solution that minimizes the trace of the covariance matrix related to relative extrinsic parameters; and (3) Pose guidance module, where optimized poses are graphically rendered to guide the acquisition of calibration images, ultimately completing the stereo calibration.

In the initial input module, after setting up the stereo system,  precise monocular calibrations are performed on the left and right cameras using a collimator \cite{bib32, bib33, bib34}, respectively. Thus, the intrinsic parameter matrices and the distortion coefficient vectors of the left and right cameras are obtained. Then, the initial stereo calibration is performed using a few (usually 2) pairs of freely captured calibration images. And the initial values of relative and absolute extrinsic parameters are calculated via PNP. These estimated parameters serve as initial input to the proposed algorithm.

In the pose output module, the reprojection residuals are computed. These residuals quantify the discrepancy between the projected image points and the observed image points. The Jacobian matrix is established by taking the partial derivatives of the reprojection residuals with respect to the estimated parameters. Additionally, the information matrix and the covariance matrix for estimating relative and absolute extrinsic parameters are constructed. A set of poses is searched in 3D space, and when the trace of the covariance matrix associated with relative extrinsic parameters reaches a threshold, this pose is referred to as the optimal pose.

In the pose guidance module, after calculating the next optimal pose, the system guides the user to complete the stereo calibration through an interactive interface. As shown in Fig. \ref{fig2}(c), the system displays the corner coordinates corresponding to the next optimal pose within the frame, and users move the calibration target to the designated position according to the pose guidance. Once the target is positioned correctly, users capture a synchronized pair of images of the calibration target at this pose. These images are added to the calibration dataset for refining the relative extrinsic parameters. Iterate the aforementioned procedures until the predetermined calibration accuracy is attained.

\subsection{Caculate Optimal Pose}\label{subsec32}

The process of calculating optimal pose is summarized in Algorithm 1. First, the intrinsic parameter matrices $\mathbf{K}^l, \mathbf{K}^r$ and the distortion coefficient vectors $\mathbf{d}^l, \mathbf{d}^r$ of the left and right cameras are calibrated using a collimator. Consequently, the optimization parameter in Eq. \eqref{equ4} can be simplified to $\mathbf{X}_i = \{ \mathbf{R}_i^l,\mathbf{t}_i^l, \mathbf{R}_i^r, \mathbf{t}_i^r \}$, where $\mathbf{R}_i^r = \mathbf{R}\mathbf{R}_i^l$ and $\mathbf{t}_i^r = \mathbf{R}\mathbf{t}_i^l + \mathbf{t}$. In a well-configured stereo measurement system, the relative extrinsic parameters between the left and right cameras are considered to be constant. The goal of our work is to make the relative extrinsic parameter estimation more accurate by adjusting the pose of the calibration target, which is named the optimal pose. 

Subsequently, several pairs of calibration images are captured, and the absolute extrinsic parameters of the left and right cameras are calculated using the PNP. On this basis, the initial value of relative extrinsic parameters can be solved using the variant of Eq. \eqref{equ3}. Except for camera parameters, both the absolute extrinsic parameters of the left camera and the relative extrinsic parameters serve as initial inputs to the algorithm.

Let us construct the Jacobian matrix $\mathbf{J}$ composed of partial derivatives of reprojection residuals with respect to relative and absolute extrinsic parameters. For notational simplicity, we define the reprojection residuals of the left and right cameras as:
\begin{equation}
\left\{
\begin{aligned}
\mathbf{\overset{\scriptscriptstyle\frown}{p}}_{ij}^l &= \mathbf{p}_{ij}^l - {\pi} (\mathbf{K}^l,\mathbf{d}^l,\mathbf{R}_i^l,\mathbf{t}_i^l,{\mathbf{P}_j}) \\
\mathbf{\overset{\scriptscriptstyle\frown}{p}}_{ij}^r &= \mathbf{p}_{ij}^r - {\pi} (\mathbf{K}^r,\mathbf{d}^r, \mathbf{R}\mathbf{R}_i^l,\mathbf{R}\mathbf{t}_i^l + \mathbf{t},{\mathbf{P}_j})
\end{aligned}
\right.
\label{equ5}
\end{equation}

\noindent where $\mathbf{\overset{\scriptscriptstyle\frown}{p}}_{ij}^l$ and $\mathbf{\overset{\scriptscriptstyle\frown}{p}}_{ij}^r$ represent the reprojection residuals of the $j$-th image point of the $i$-th image, respectively. Inspired by Peng et al. \cite{bib28},  we derive the Jacobian matrices of the reprojection residuals as follows:
\begin{equation}
\mathbf{U}_i = \left( \begin{array}{cc}
\frac{\partial \mathbf{\buildrel{\lower3pt\hbox{$\scriptscriptstyle\frown$}} \over p}_{i1}^r}{\partial \mathbf{R}} & \frac{\partial \mathbf{\buildrel{\lower3pt\hbox{$\scriptscriptstyle\frown$}} \over p}_{i1}^r}{\partial \mathbf{t}} \\
\vdots & \vdots \\
\frac{\partial \mathbf{\buildrel{\lower3pt\hbox{$\scriptscriptstyle\frown$}} \over p}_{in}^r}{\partial \mathbf{R}} & \frac{\partial \mathbf{\buildrel{\lower3pt\hbox{$\scriptscriptstyle\frown$}} \over p}_{in}^r}{\partial \mathbf{t}}
\end{array} \right)
\mathbf{V}_i = \left( \begin{array}{cc}
\frac{\partial \mathbf{\buildrel{\lower3pt\hbox{$\scriptscriptstyle\frown$}} \over p}_{i1}^l}{\partial \mathbf{R}_i^l} & \frac{\partial \mathbf{\buildrel{\lower3pt\hbox{$\scriptscriptstyle\frown$}} \over p}_{i1}^l}{\partial \mathbf{t}_i^l} \\
\vdots & \vdots \\
\frac{\partial \mathbf{\buildrel{\lower3pt\hbox{$\scriptscriptstyle\frown$}} \over p}_{in}^l}{\partial \mathbf{R}_i^l} & \frac{\partial \mathbf{\buildrel{\lower3pt\hbox{$\scriptscriptstyle\frown$}} \over p}_{in}^l}{\partial \mathbf{t}_i^l}
\end{array} \right)
\label{equ6}
\end{equation}

Specifically, $\mathbf{U}_i$ denotes the right camera's reprojection residuals with respect to the relative extrinsic parameters, while $\mathbf{V}_i$ represents the left camera's reprojection residuals with respect to its absolute extrinsic parameters. The explicit expressions of matrix blocks $\mathbf{U}_i$ and $\mathbf{V}_i$ are provided in Appendix \ref{secA}.

Based on this formulation, we construct a highly sparse Jacobian matrix that simultaneously encodes the relationships between the absolute extrinsic parameters of the left camera and the relative extrinsic parameters:
\begin{equation}
\mathbf{J} = \left( {\begin{array}{*{20}{c}}
\mathbf{U}_1 & \mathbf{V}_1 & \mathbf{0} & \cdots & \mathbf{0} \\
\mathbf{U}_2 & \mathbf{0} & \mathbf{V}_2 & \cdots & \mathbf{0} \\
\vdots & \vdots & \vdots & \ddots & \vdots \\
\mathbf{U}_m & \mathbf{0} & \mathbf{0} & \cdots & \mathbf{V}_m
\end{array}} \right)
\label{equ7}
\end{equation}

Next, establish the information matrix $\mathbf{J}^T \mathbf {J}$, which can be expressed by block matrix as:
\begin{equation}
\mathbf{J}^T\mathbf{J} 
= \begin{pmatrix}
\sum\limits_{i = 1}^m \mathbf{U}_i^T\mathbf{U}_i & \mathbf{U}_1^T\mathbf{V}_1 & \mathbf{U}_2^T\mathbf{V}_2 & \cdots & \mathbf{U}_m^T\mathbf{V}_m \\
\mathbf{V}_1^T\mathbf{U}_1 & \mathbf{V}_1^T\mathbf{V}_1 & \mathbf{0} & \cdots & \mathbf{0} \\
\mathbf{V}_2^T\mathbf{U}_2 & \mathbf{0} & \mathbf{V}_2^T\mathbf{V}_2 & \cdots & \mathbf{0} \\
\vdots & \vdots & \vdots & \ddots & \vdots \\
\mathbf{V}_m^T\mathbf{U}_m & \mathbf{0} & \mathbf{0} & \cdots & \mathbf{V}_m^T\mathbf{V}_m
\end{pmatrix}
= \begin{pmatrix}
\mathbf{A} & \mathbf{C} \\
\mathbf{C}^T & \mathbf{B}
\end{pmatrix}
\label{equ8}
\end{equation}

\noindent with
\begin{equation}
\left\{ 
\begin{aligned}
\mathbf{A} &= \sum\limits_i \mathbf{U}_i^T\mathbf{U}_i \\
\mathbf{B} &= \text{diag}(\mathbf{V}_1^T\mathbf{V}_1, \cdots ,\mathbf{V}_m^T\mathbf{V}_m) \\
\mathbf{C} &= (\mathbf{U}_1^T\mathbf{V}_1, \cdots ,\mathbf{U}_m^T\mathbf{V}_m)
\end{aligned} 
\right.
\label{equ9}
\end{equation}

The inverse of the information matrix $({\mathbf{J}^T \mathbf {J}})^{-1}$ provides an estimate of the covariance matrix for both the absolute and relative extrinsic parameters. In particular, the upper-left submatrix of $(\mathbf{J}^T \mathbf{J})^{-1}$ can be used to estimate the covariance of relative extrinsic parameters, which is expressed as $\text{Cov}\left( \theta_{\text{rel}} \right) = \left( \mathbf{A} - \mathbf{C}\mathbf{B}^{-1}\mathbf{C}^T \right)^{-1} = \mathbf{\Sigma}$. Relevant proofs are provided in Appendix \ref{secB}.

\begin{algorithm}[t]
\caption{Optimal pose calculation algorithm}
\label{alg:pose_calc}
\textbf{Input:} \\
$\{\mathbf{K}^l, \mathbf{K}^r\}$ -- camera intrinsic matrix; $\{\mathbf{d}^l, \mathbf{d}^r\}$ -- distortion coefficients. \\
\textbf{Output:} \\
$\mathbf{R}^l$ -- optimal rotation vector and $\mathbf{t}^l$ -- optimal translation vector. 
\begin{algorithmic}[1]
    \State Capture $m$ pairs of calibration images to compute $\{\mathbf{R}_m^l, \mathbf{t}_m^l, \mathbf{R}_m^r, \mathbf{t}_m^r\}$ via PnP.
    \State Calculate the initial value of $\{\mathbf{R}, \mathbf{t}\}$.
    \For{$i = 1$ \textbf{to} $m$}
    \State Calculate the Jacobian matrix $\mathbf{J}_m$ using Eq.\eqref{equ5} -- Eq.\eqref{equ7}.
    \EndFor
    \While{$\Delta RE > 1\mathrm{e}{-6}$ \textbf{and} $\text{iterations} < 500$}
        \State Generate candidate pose $\{\mathbf{R}_{m+1}^l, \mathbf{t}_{m+1}^l\}$.
        \If{The target in the candidate pose is in the common FOV}
            \State Update the Jacobian matrix to $\mathbf{J}_{m+1}$ using Eq.\eqref{equ5} -- Eq.\eqref{equ7}.
            \State Calculate the trace of the covariance matrix $\mathrm{tr}(\boldsymbol{\Sigma}_{m+1})$.
            \State Calculate the optimal pose using Eq.\eqref{equ10}.
        \EndIf
\EndWhile
    \State Return next optimal pose $\mathbf{R}_{m+1}^l, \mathbf{t}_{m+1}^l$.
\end{algorithmic}
\end{algorithm}

From this, we proceed to compute the next optimal pose. Specifically, we first establish a set of absolute extrinsic parameters of the left camera in 3D space (i.e., the pose of the calibration board relative to the left camera) and set the initial value of the relative extrinsic parameters as the ground truth. When a calibration board corresponding to a specific absolute extrinsic parameter lies within the common FOV, this absolute extrinsic parameter is regarded as a candidate for the next optimal pose. This optimization problem can be formulated as:
\begin{equation}
\mathbf{T}_{m + 1}^* = \arg \min_{\mathbf{T}_{m + 1}} \text{tr}\left( \mathbf{\Sigma}_{m + 1} \right)
\label{equ10}
\end{equation}

The above process is executed iteratively, with termination conditions: the number of iterations reaches or exceeds 500, or the relative error of the trace is less than or equal to $10^{-6}$. Thus, we obtain the next optimal pose. Subsequently, a visualization-based guidance interface is employed to assist the user in capturing calibration images. Ultimately, the bundle adjustment approach is applied to jointly optimize the system parameters and the image coordinates, thereby completing the stereo calibration of the binocular camera system.

\section{Experiments and Application}\label{sec4}

We have compared our method with independent monocular calibration and stereo-specialized calibration methods \cite{bib17}. The experimental procedures and results are presented in Appendix \ref{secC} and Appendix \ref{secD}, respectively. And to quantitatively evaluate the effectiveness of the proposed method, we conducted comparative experiments between the optimal pose derived from our optimization framework and the constrained random pose. Note that the random pose is not completely random, which is carefully designed to avoid degenerate configurations. Inspired by \cite{bib27, bib21, bib22}, the random pose follows the following principles:

\begin{enumerate}[label=\textbullet]
    \item \textbf{Full-FOV coverage constraint}: Throughout the calibration sequence, the composite projection of all calibration target positions must cover more than 90\% of the overlapping FOV between both cameras to ensure complete parameter observability.
     \item \textbf{Optical axis non-alignment constraint}: The calibration target's surface normal vector must not be parallel to the optical axes of either camera to ensure adequate depth observability and prevent degenerate configurations in stereo vision.
      \item \textbf{Sampling constraint}: Random poses are sampled within the SE(3) space, where rotational parameters follow a random distribution within ±30°, and the calibration target is randomly moved within the overlapping FOV. 
\end{enumerate}

\subsection{Simulation Evaluations}
\label{subsec41}

To verify the robustness of the method proposed in this paper, Gaussian noise with different standard deviations was introduced in the simulation experiments. The experimental hardware platform used in this paper is configured with an NVIDIA GeForce RTX 3050 Laptop GPU, featuring a core frequency of 3.3 GHz and 16 GB of RAM. Here, we briefly introduce the simulation experiment settings.

Specifically, the calibration target adopted in the experiment is a $9 \times 6$ 2D chessboard with a grid spacing of 5 mm, and the origin of the world coordinate system is set on the chessboard plane. Based on the aforementioned principles, a set of random poses for the absolute extrinsic parameters of the left camera is defined, and the analysis is conducted by comparing the calibration results of the optimal pose and random poses. In addition, the parameters of the left and right cameras are consistent, with specific parameters as follows: resolution of $640 \times 480$ pixels, principal point coordinates of $(320,240)$, focal length of $800$, and distortion coefficients of $ \left[ 0.01, 0.1, 0, 0 \right]^T $, the relative extrinsic parameters are $ \mathbf{R} = \left[ -0.003, -0.303, -0.017 \right]^T$ and $\mathbf{t} = \left[ 440.3, -6.2, 25.1 \right]^T$.

The experimental process is as follows:

\begin{figure}[t]
    \centering
    \includegraphics[width=1\textwidth]{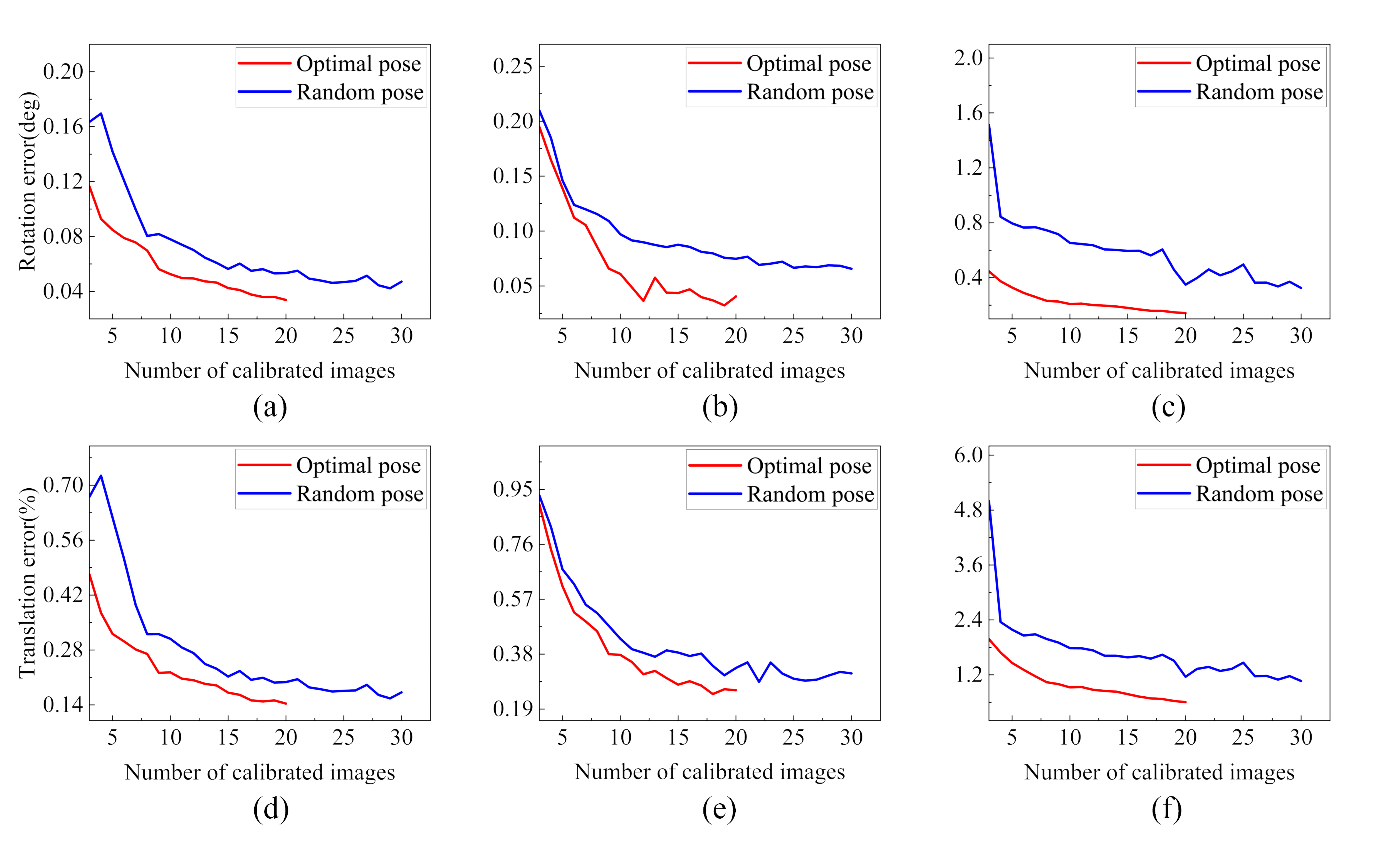}
    \caption{Comparison of relative extrinsic parameter estimation errors between random poses and optimized poses under Gaussian noise with different standard deviations. (a) Rotation error, and (d) translation error under Gaussian noise with a standard deviation of 0.5 pixel; (b) rotation error, and (e) translation error under Gaussian noise with a standard deviation of 1 pixel; (c) rotation error, and (f) translation error under Gaussian noise with a standard deviation of 2 pixel}
    \label{fig3}
\end{figure}

First, two pairs of random images in the common FOV are generated. After adding Gaussian noise to the image points, the initial values of the relative extrinsic parameters are solved. These two pairs of random images serve as calibration images for both the random pose method and the optimized pose method. Subsequently, $28$ images are generated according to the definition of random poses, resulting in a total of $28+2=30$ calibration images. Meanwhile, $18$ images are generated through optimal pose, yielding a total of $18+2=20$ calibration images. For Gaussian noise with the same standard deviation in the same group of experiments, three standard deviation levels are set, that is, $0.5$ pixel, $1$ pixel, and $2$ pixels. The relative extrinsic parameters are solved by inputting the calibration images. $100$ trials are performed for each experiment. 

To evaluate the calibration results, the calculated relative extrinsic parameters are compared with the ground truth for error assessment, and the specific error calculation formulas are as follows:
\begin{equation}
E(\mathbf{R}_{\text{cal}}) = \arccos\left( \frac{\text{tr}(\mathbf{R}_{\text{ref}} \mathbf{R}_{\text{cal}}^\top) - 1}{2} \right) \cdot \frac{180}{\pi} 
\end{equation}

\begin{equation}
E(\mathbf{t}_{\text{cal}}) = 2 \left( \frac{\|\mathbf{t}_{\text{ref}} - \mathbf{t}_{\text{cal}}\|}{\|\mathbf{t}_{\text{ref}}\| + \|\mathbf{t}_{\text{cal}}\|} \right) \cdot 100\%
\end{equation}

\noindent where $\left( \mathbf{R}_{\text{ref}}, \, \mathbf{t}_{\text{ref}} \right)$ denotes the ground truth, and $\left( \mathbf{R}_{\text{cal}}, \, \mathbf{t}_{\text{cal}} \right)$ represents the calculated result. As illustrated in Fig. \ref{fig3}, it can be observed that, when an appropriate number of calibration images are employed, the calculated value of the relative extrinsic parameter gradually converges to the ground truth with the increase in the number of calibration images. In comparison with the random pose strategy, the relative extrinsic parameter calibrated via the optimal pose exhibits a higher consistency with the ground truth and converges at a faster rate, thus validating the effectiveness of the proposed method. Additionally, when Gaussian noise with standard deviations of $0.5$ pixel, $1$ pixel, and $2$ pixels is introduced, the calibration result based on the optimal pose still maintains a rapid convergence speed, further corroborating the robustness of the proposed method.

\subsection{Comparative Experiment on Calibration Efficiency}
\label{subsec42}

The experimental procedure was designed as follows. Initially, both cameras were rigidly mounted on a stereoscopic rig with adjustable baselines to form a stereo vision system, ensuring that they had an appropriate overlapping FOV. And individual calibration of the left and right cameras was conducted to obtain precise intrinsic parameters and distortion coefficients. Subsequently, an operator manipulated the calibration target in accordance with the above principles of random pose and captured 20 calibration images. We conducted stereo calibration with selected sets of image pairs (e.g., 2, 10, and 20 pairs). And the stereo calibration was completed through the optimization of the minimal reprojection error, which is widely recognized and employed in practical applications. To ensure comparability, we selected the first two images captured with the aforementioned random poses as the initial input for calculating the optimal pose, and then captured the calibration images under the guidance of our proposed method until the reprojection error reached that of random-pose-based calibration. 

\begin{figure}[htbp]
\centering
\includegraphics[width=0.9\textwidth]{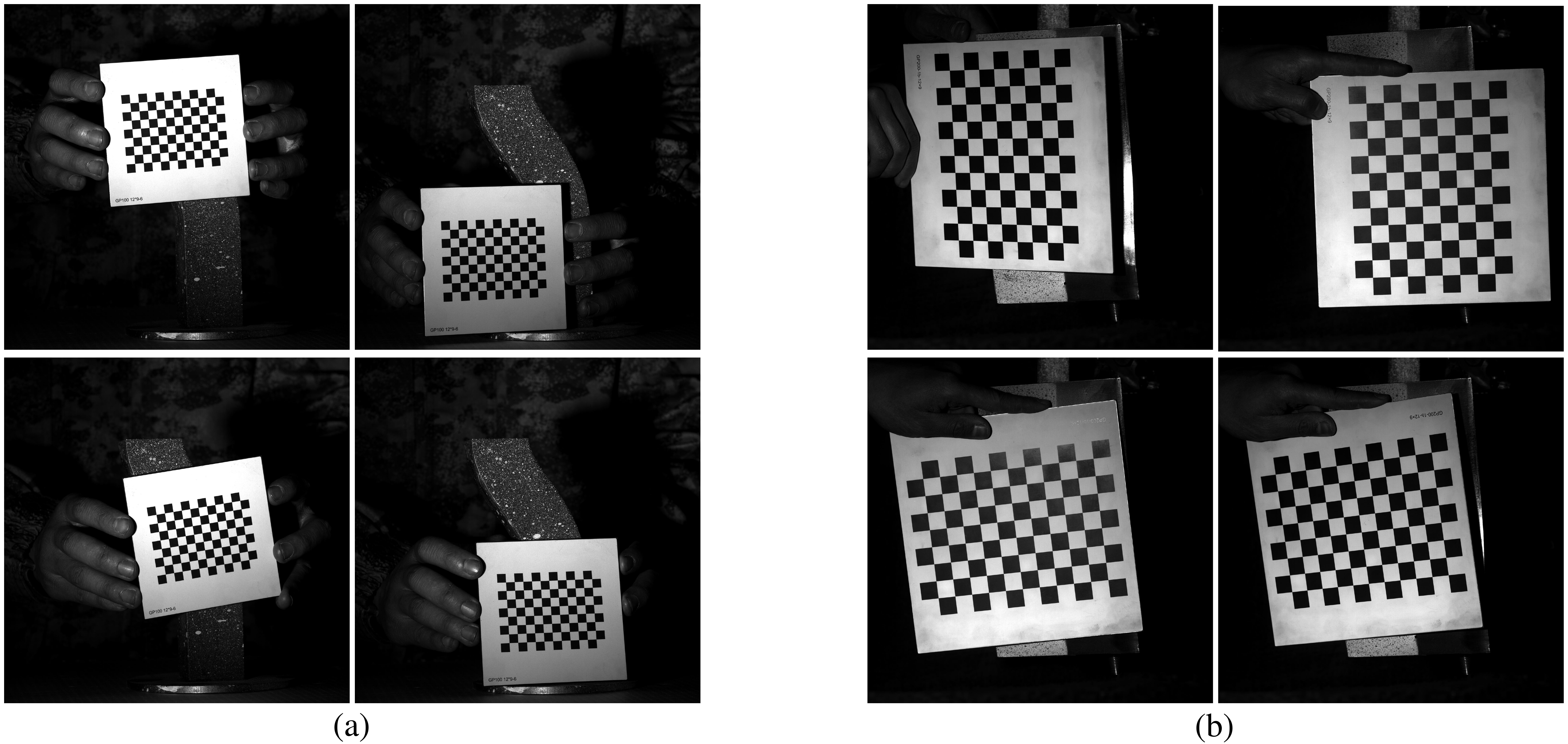}
\caption{Sample calibration images in the comparative experiments. \textbf{(a)} The calibration target approximately occupied $ 1/9 $ of the common FOV, and \textbf{(b)} the calibration target approximately occupied $ 1/2 $ of the common FOV}\label{fig4}
\end{figure}

Furthermore, to verify the robustness of the proposed method, two distinct experimental conditions were employed to conduct the comparative experiments. The configuration of the stereo vision system varied with different experimental conditions, while the parameters of the two cameras were approximately the same, with the FOVs of the two systems roughly between 20° and 40°. As shown in Fig. \ref{fig4}(a), the difference in the comparative experiments lay in the proportion of the calibration target within the common FOV, which was used to test the performance of the proposed method.  In the low-coverage case ($l$-system), the calibration target approximately occupied $ 1/9 $ of the overlapping FOV in terms of area (Fig. \ref{fig4}(a)). And in the high-coverage case ($h$-system), the calibration target occupied approximately $ 1/2 $ of the overlapping FOV in terms of area( Fig. \ref{fig4}(b)).

\begin{table}[h]
\caption{The experimental results for the $l$ system}\label{tab1}
\begin{tabular*}{\textwidth}{@{\extracolsep\fill}c@{\extracolsep\fill}cccc}
\toprule%
\multicolumn{1}{c}{\textbf{Calibration scheme}} & \textbf{R}/deg & \textbf{t}/mm & \textbf{RE}/pixel  & \textbf{TE}/mm  \\ 
\midrule
2-random          & $(-1.350, -18.698, -0.515)$ & $(-376.5, 12.8, 67.7)$ & 0.6179   & 0.0973   \\
10-random         & $(-0.592, -19.528, -0.517)$  & $(-390.9, -1.9, 75.0)$ & 0.3587   & 0.0894   \\
20-random         & $(-0.382, -19.465, -0.503)$  & $(-385.0, -5.8, 72.0)$ & 0.2808   & 0.0739  \\
2-random + 2-optimal & $(-0.969, -18.960, -0.512)$  & $( -381.2, 5.4, 69.0)$ & 0.4150   & 0.0846   \\
2-random + 4-optimal & $(-0.474, -19.127, -0.526)$  & $(-384.0, -4.1, 71.3)$ & 0.3323   & 0.0690   \\
2-random + 6-optimal & $(-0.388, -19.186, -0.532)$  & $(-385.0, -5.8, 72.0)$ & 0.2715   & 0.0676   \\
\botrule
\end{tabular*}
\footnotetext{Note: The table presents reprojection errors (\textbf{RE}) in pixels, and triangulation errors (\textbf{TE}) in millimeters for different calibration schemes. For clarity, \textbf{R} in this paper is expressed in vector}
\end{table}

To comprehensively evaluate the calibration accuracy, we employed both the reprojection error and the triangulation error as quantitative metrics. The reprojection error was determined by calculating the Euclidean distance between the image point coordinates predicted by the projection equations and the actual extracted image point coordinates. The triangulation error was calculated by performing spatial intersection using corresponding pixel points from the left and right image planes, based on the calibration results, to compute the 3D point coordinates. These computed 3D coordinates were then compared with the 3D coordinates of the calibration target by calculating their Euclidean distance. This dual-metric approach provides complementary perspectives on calibration accuracy, assessing both the 2D image plane consistency and 3D spatial reconstruction fidelity. 

Table \ref{tab1} presents the experimental results when the calibration target occupies $ 1/9 $ of the FOV. It demonstrates that the optimal pose significantly outperforms random pose. When comparing 10-random with 2-random + 4-optimal configurations in the l-system, the proposed approach reduced reprojection error by 7.4\% (from 0.3587 to 0.3323 pixels) and triangulation error by 22.8\% (from 0.0894 to 0.0690 mm). The comparison between 20-random and 2-random + 6-optimal configurations showed slightly diminished but still meaningful gains, with reprojection error improving by 3.3\% (from 0.2808 to 0.2715 pixels) and triangulation error by 8.5\% (from 0.0739 to 0.0676 mm) in the l-system.

\begin{table}[h]
\caption{The experimental results for the $h$ system}\label{tab2}
\begin{tabular*}{\textwidth}{@{\extracolsep\fill}c@{\extracolsep\fill}cccc}
\toprule%
\multicolumn{1}{c}{\textbf{Calibration scheme}} & \textbf{R}/deg & \textbf{t}/mm & \textbf{RE}/pixel & \textbf{TE}/mm \\ 
\midrule
2-random          & $(-0.069, 17.578, 1.073)$ & $(438.8, -3.5, 31.2)$ & 0.1582   & 0.0551    \\
10-random         & $(0.038, 17.587, 1.061)$  & $(438.9, -6.0, 31.2)$ & 0.1417  & 0.0487   \\
20-random         & $(0.061, 17.621, 1.051)$  & $(439.7, -6.5, 31.6)$ & 0.1222   &  0.0455   \\
2-random + 2-optimal & $(-0.002, 17.632, 1.059)$  & $(440.0, -5.0, 31.6)$ & 0.1194   & 0.0476  \\
2-random + 4-optimal & $(0.047, 17.614, 1.059)$  & $(439.5, -6.2, 31.6)$ &  0.1239  & 0.0474  \\
2-random + 6-optimal & $(0.049, 17.615, 1.056)$  & $(439.6, -6.3, 31.6)$ & 0.1215   & 0.0473  \\
\botrule
\end{tabular*}
\end{table}

Similarly, Table \ref{tab2} presents the experimental results when the calibration target occupies $ 1/2 $ of the FOV.  The data demonstrate that the 2-random + 2-optimal configuration achieves convergence, with the reprojection error reduced to 0.1194 pixels, which is superior to the 20-random approach (0.1222 pixels). Error distribution analysis confirms this convergence state, showing that subsequent increases to 4 or 6 optimized poses produce only minimal variations in both reprojection error (within ± 0.0025 pixels) and triangulation error (within ± 0.0003 mm). 

As evidenced by Table \ref{tab1} and Table \ref{tab2}, the optimized pose achieves comparable calibration accuracy to the random pose, while simultaneously reducing the required number of calibration images. Additionally, the method maintains robustness across different experimental conditions, demonstrating its practical applicability.

\subsection{Comparative Experiment on Calibration Accuracy}
\label{subsec43}

\begin{figure}[t]
\centering
\includegraphics[width=0.6\textwidth]{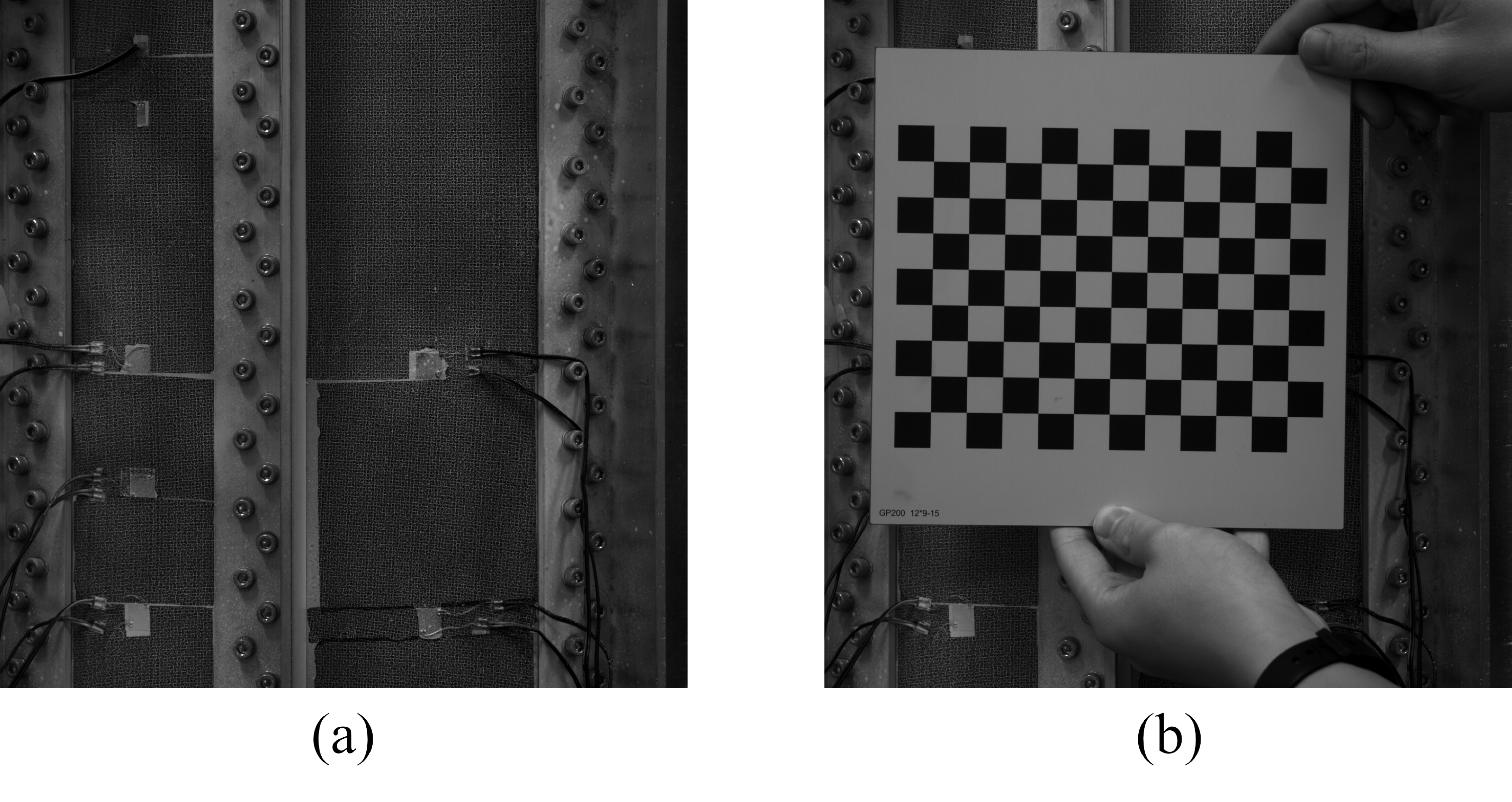}
\caption{Comparative experiment on calibration accuracy. \textbf{(a)} Experimental setup, and \textbf{(b)} sample of the calibration images}\label{fig5}
\end{figure}

Furthermore, we quantitatively compared the calibration accuracy between the random pose and the optimal pose using the same number of calibration images, and evaluated the calibration results based on strain measurement data. Figure \ref{fig5}(a) shows the experimental setup, a cylindrical specimen is prepared with randomly distributed water-transfer speckle patterns and strategically mounted strain gauges, whose measurements served as the ground truth for strain verification. Figure \ref{fig5}(b) presents a sample of the calibration images, where the checkerboard occupies approximately $ 1/2 $ of the FOV. We established a rigid stereo vision system through two IDS industrial cameras (resolution: 2048×2048 pixels). Based on the data in Table \ref{tab2}, we compared the stereo calibration results between the 8-random configuration and the 2-random + 6-optimal configuration. The results of stereo calibration are presented in Table \ref{tab3}.

\begin{figure}[t]
\centering
\includegraphics[width=1\textwidth]{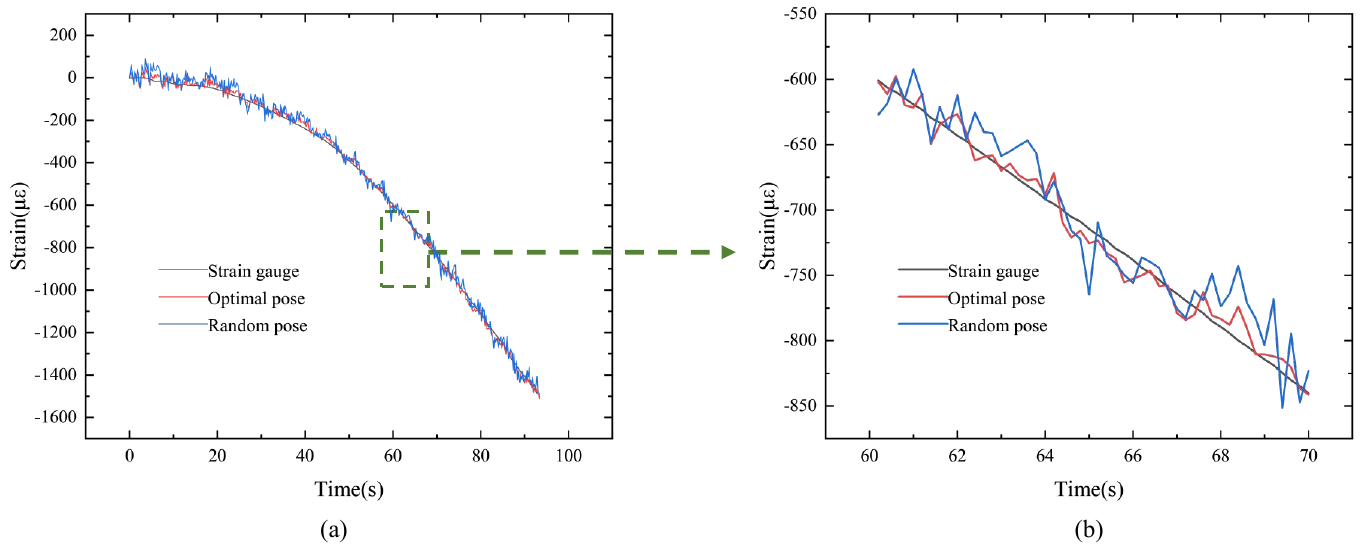}
\caption{Comparative analysis of ${\varepsilon_{yy}}$ strain distributions. \textbf{(a)} Macroscopic strain distribution under full loading history, and \textbf{(b)} local strain concentration}
\label{fig6}
\end{figure}

The experiment conducted was an axial compression test, where a constant loading rate was applied to the cylindrical specimen until reaching the target load of 80 metric tons. Throughout the process, the stereo vision system captured deformation images of the specimen at 5 fps. The stereo calibration results obtained from random poses and optimal poses were integrated with DIC software to compute the specimen's strain fields, respectively. After synchronizing the sampling frequencies of the strain measurement device and the stereo vision system, we performed the analysis of their strain data.

\begin{table}[t]
\caption{The experimental results of stereo calibration}\label{tab3}
\begin{tabular*}{\textwidth}{@{\extracolsep\fill}c@{\extracolsep\fill}cccc}
\toprule%
\multicolumn{1}{c}{\textbf{Calibration scheme}} & \textbf{R}/deg & \textbf{t}/mm & \textbf{RE}/pixel\\ 
\midrule
8-random                    & $(0.407, -17.614, -1.118) $& $(439.3 -4.5, 38.2) $ & 0.0531        \\
2-random + 6-optimal        & $(0.601, -18.164, -1.151)$ & $(438.2, -1.1, 48.4) $ & 0.0471       \\
\botrule
\end{tabular*}
\end{table}

\begin{figure}[t]
\centering
\includegraphics[width=0.5\textwidth]{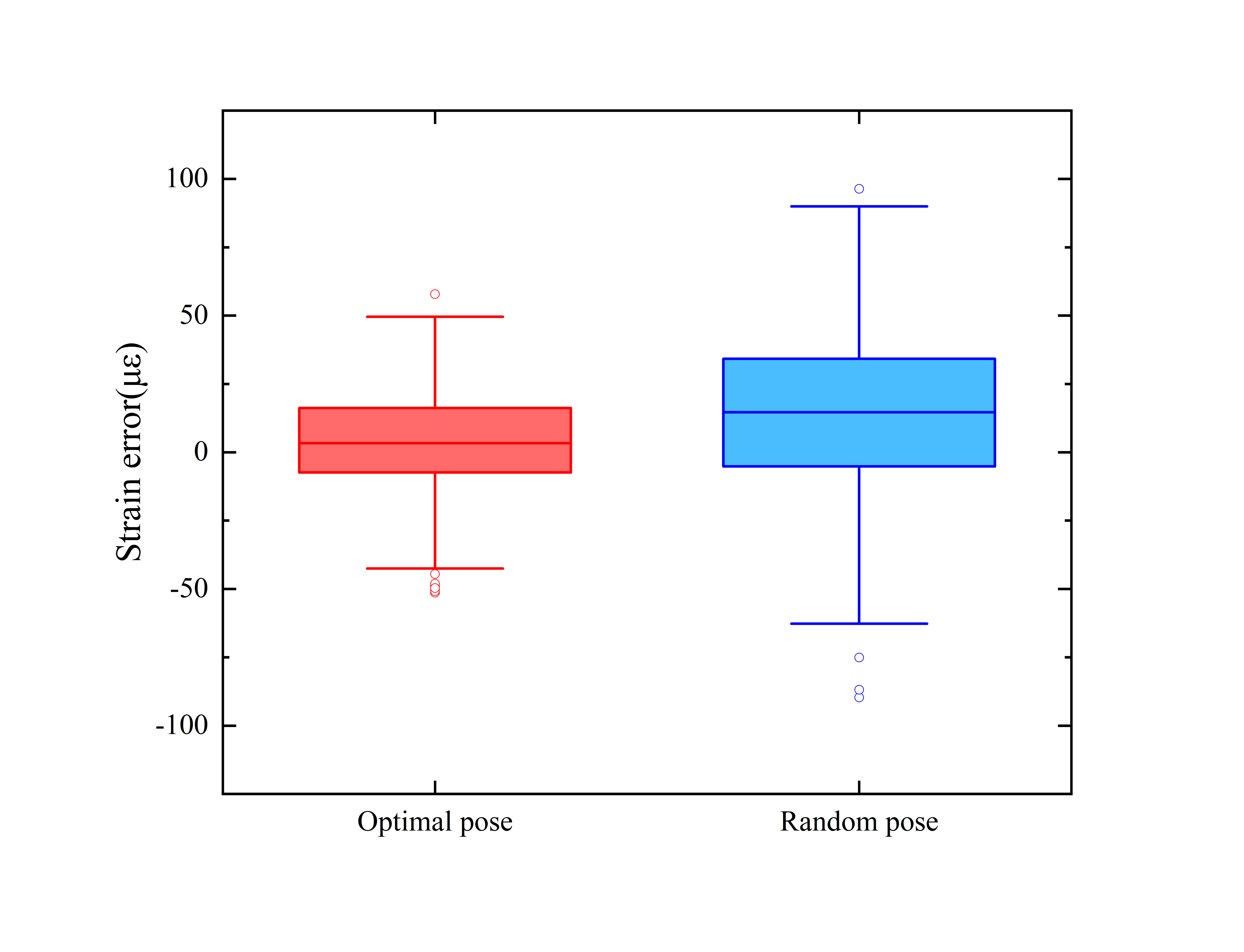}
\caption{Box diagram of longitudinal strain ${\varepsilon_{yy}}$}
\label{fig7}
\end{figure}

Figure \ref{fig6} presents the longitudinal strain ${\varepsilon_{yy}}$ of the specimen. Figure \ref{fig6}(a) shows the comparative strain diagram during the entire loading process, while Fig. \ref{fig6}(b) provides a segmental strain comparison for enhanced visualization. As shown in Fig. \ref{fig6}(b), a qualitative analysis reveals that, using strain gauge measurements as ground truth, the optimal-pose calibration yields strain values that are closer to the ground truth with less variability compared to random-pose calibration.

As shown in Fig. \ref{fig7}, post-processing was applied to the strain data. The strain error, defined as the differences between the calculated values (obtained from both optimal and random pose calibrations) and the strain gauge measurements, was computed and visualized using a box diagram. The strain error between the calculated values from optimal pose calibration and strain gauge measurements had a mean of 3.3 $~\mu\varepsilon$ with a standard deviation of 18.2 $~\mu\varepsilon$, whereas the residuals from random pose calibration showed a mean of 14.7 $~\mu\varepsilon$ and a standard deviation of 30.2 $~\mu\varepsilon$. The box diagram and local strain analysis further confirmed that optimal pose reduces measurement variance and enhances strain field reconstruction accuracy.

\subsection{Application of Thermal Deformation Measurement}
\label{subsec44}

Investigating the deformation characteristics of the S-shaped inlet duct of a hypersonic vehicle under high-temperature conditions is of significant importance for its rational design. However, constrained by current experimental conditions (e.g., limited operating space), operators must complete calibration tasks rapidly with a minimal set of calibration images. To address this, we apply the proposed method to this task. We combined the proposed method with DIC to measure the in-plane displacement of the specimen under high-temperature conditions and compared it with the numerical simulation results.

An S-shaped specimen was designed in this experiment. The geometry of the S-shaped specimen was identical to that of the S-shaped bend in the constant cross-sectional intake duct of hypersonic vehicles. A rigid stereo vision system was established by mounting two IDS industrial cameras (resolution: $2048 \times 2048$ pixels) on a stabilized tripod. The thermal deformation measurement experiment was conducted using a heating cabinet, with heating controlled by thermocouples. To mitigate thermal radiation at high temperatures, a high-power blue light source was employed, and narrow bandpass filters with a central wavelength of 450 ± 10 nm were installed on the left and right cameras. To overcome the limitations of conventional speckle patterns in high-temperature environments, we applied a specialized high-temperature-resistant speckle pattern using thermally stable ceramic-based spray paint, ensuring reliable pattern stability for accurate DIC measurements throughout the thermal loading cycle.

We completed the stereo calibration using the 2-random + 5-optimal poses and implemented the thermal deformation measurement task in conjunction with DIC software. The calibration results of the stereo vision system are shown in Table \ref{tab4}. $(f_u, f_v)$ represents the focal length, $(u_0, v_0)$ refers to the principal point, which are all camera intrinsic parameters, and the mean reprojection error is merely 0.1011 pixels.

\begin{figure}[t]
\centering
\includegraphics[width=0.8\textwidth]{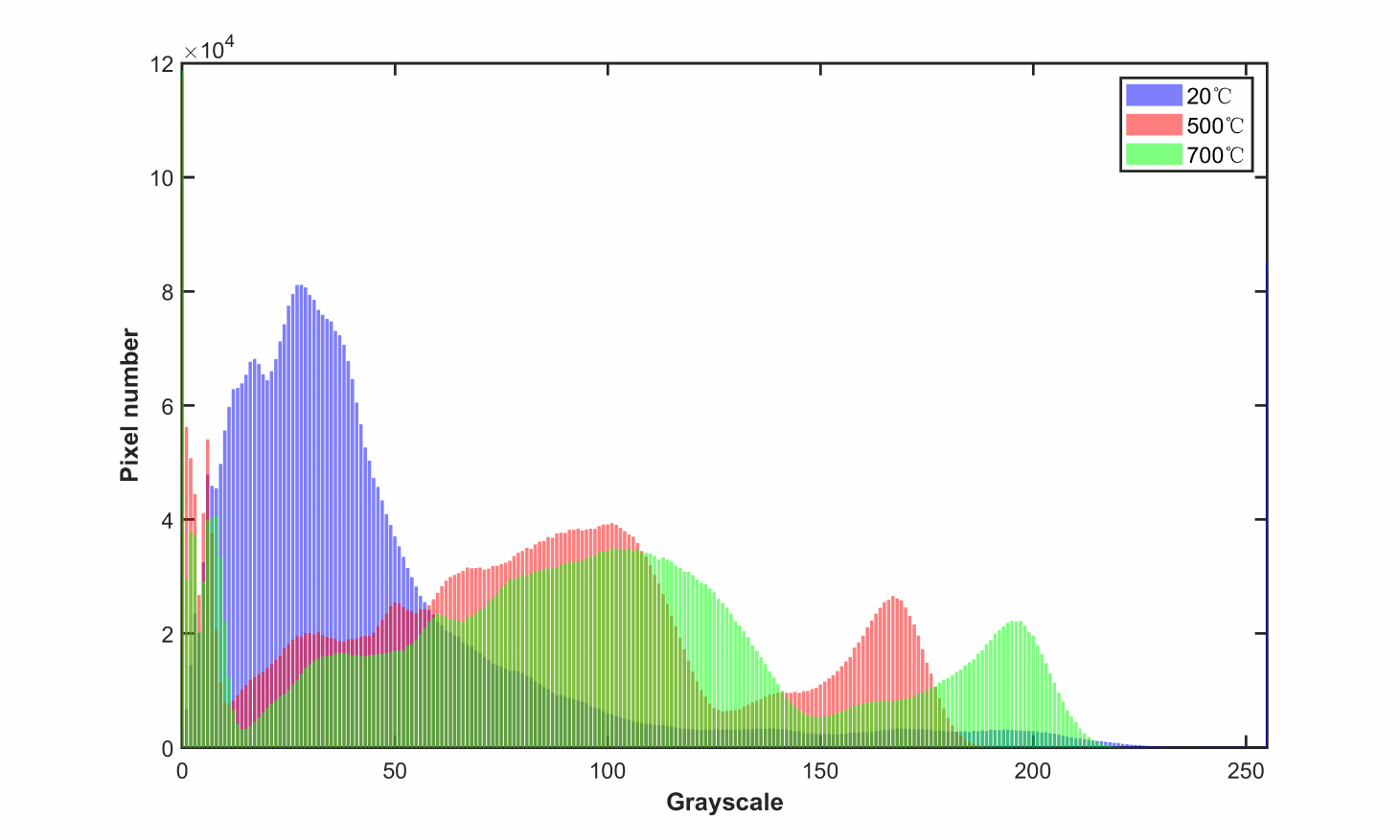}
\caption{Comparison of grayscale histograms of speckle patterns at 20°C, 500°C and 700°C}
 \label{fig8}
\end{figure}

\begin{table}[t]
\caption{Stereo camera calibration results}\label{tab4}
\begin{tabular*}{\textwidth}{@{\extracolsep{\fill}}c@{\extracolsep{\fill}}c@{\extracolsep{\fill}}c@{}}
\toprule
\multicolumn{1}{c}{Parameter} & \multicolumn{1}{c}{Left camera} & \multicolumn{1}{c}{Right camera} \\
\midrule
$(f_u, f_v)$/pixel & (10221.8, 10191.7) & (10063.6, 10079.5) \\
$(u_0, v_0)$/pixel & (965.8, 974.0) & (904.3, 943.1) \\
$(k_1, k_2)$/pixel & (0.330, 30.356 & (0.403, -0.313) \\
$(p_1, p_2)$/pixel & (-0.012, 0.002) & (-0.012, 0.006) \\
\multicolumn{1}{c}{$\textbf{R}$/rad} & \multicolumn{2}{c}{$(-0.174, 8.314, 1.023)$} \\
\multicolumn{1}{c}{$\textbf{t}$/mm} & \multicolumn{2}{c}{$(-80.0, 1.1, 9.4)$} \\
\multicolumn{1}{c}{RE/pixel} & \multicolumn{2}{c}{0.1011} \\
\botrule
\end{tabular*}
\end{table}

In this experiment, two target temperatures were set as 500°C and 700°C, respectively. To ensure that the measurement results accurately reflect the deformation at each temperature, once the temperature first reached the set point, it was maintained at that level for 20 minutes, during which the camera continuously captured images. For deformation analysis, we selected the frame captured at the 20-minute mark as the reference image for DIC computation, representing the steady-state thermal deformation at each target temperature.

To verify the stability of speckle patterns under the high-temperature conditions established in this study, we evaluated the speckle patterns at 20°C, 500°C, and 700°C using three key metrics \cite{bib36, bib37}. Fig. \ref{fig8} shows the distribution of pixel numbers across various grayscale values for speckle patterns captured at 20°C, 500°C, and 700°C. The histograms reveal a shift in brightness and the presence of distinct peaks at low and high grayscale ranges, indicating good contrast and texture information suitable for DIC analysis. To further evaluate image quality, we analyze the mean intensity (MI) and mean intensity gradient (MIG). As shown in Table \ref{tab5}, MI increases with temperature due to thermal radiation, yet no pixel saturation or overexposure is observed. MIG values remain comparable across all temperatures and slightly increase at 700°C, indicating maintained image quality and suitability for DIC.

\begin{figure}[t]
\centering
\includegraphics[width=0.8\textwidth]{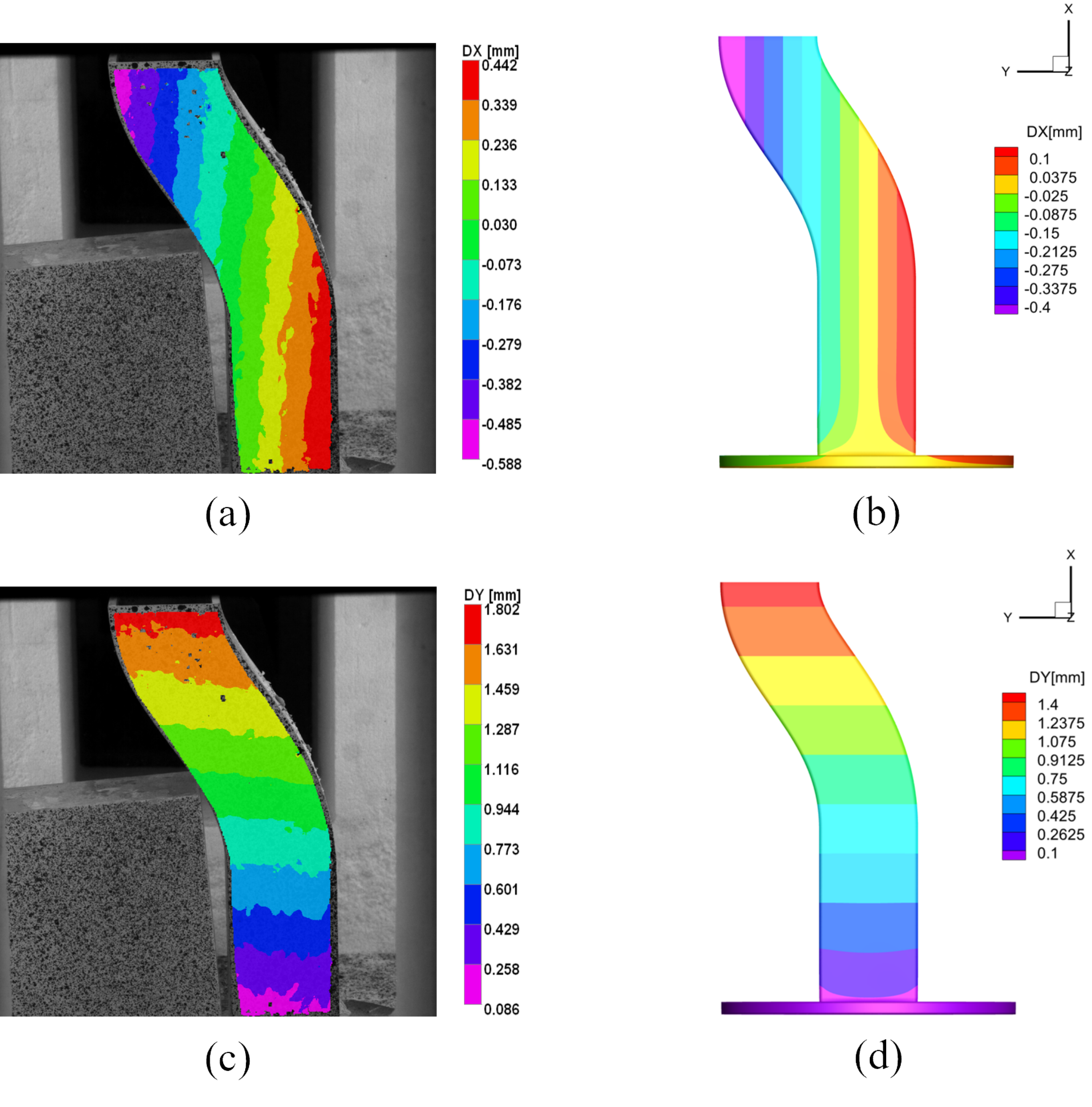}
\caption{Thermal deformation measurement results at 500℃. \textbf{(a)} Actual displacement U, \textbf{(b)} simulated displacement U, \textbf{(c)} actual displacement V, and \textbf{(d)} simulated displacement V}
 \label{fig9}
\end{figure}

\begin{table}[t]
\centering
\caption{Comparative analysis of MI and MIG at 20°C, 500°C and 700°C}
\label{tab5}
\begin{tabular*}{\textwidth}{@{\extracolsep{\fill}}c@{\extracolsep{\fill}}c@{\extracolsep{\fill}}c@{\extracolsep{\fill}}c@{}}
\hline
Statistical parameters & $20^\circ\text{C}$ & $500^\circ\text{C}$ & $700^\circ\text{C}$ \\
\hline
MI  & 50.43 & 81.48 & 95.38 \\
MIG & 6.45  & 6.85  & 6.87  \\
\hline
\end{tabular*}
\end{table}

To validate the measurement results, we computed the displacement field of the specimen using the open-source FEA software MOOSE, with post-processing visualization performed in Tecplot 360. The S-bend specimen was fabricated from 310S stainless steel, with temperature-dependent material properties carefully selected for the simulation: Young’s modulus values are $163$ GPa and $146$ GPa, and the coefficients of thermal expansion are $17.0 \times10^{-6}$~℃$^{-1}$ and $18.0 \times10^{-6}$~℃$^{-1}$ at 500℃ and 700℃, respectively. These values, along with all other material parameters, were sourced from established literature \cite{bib35}. 

\begin{figure}[t]
\centering
\includegraphics[width=0.8\textwidth]{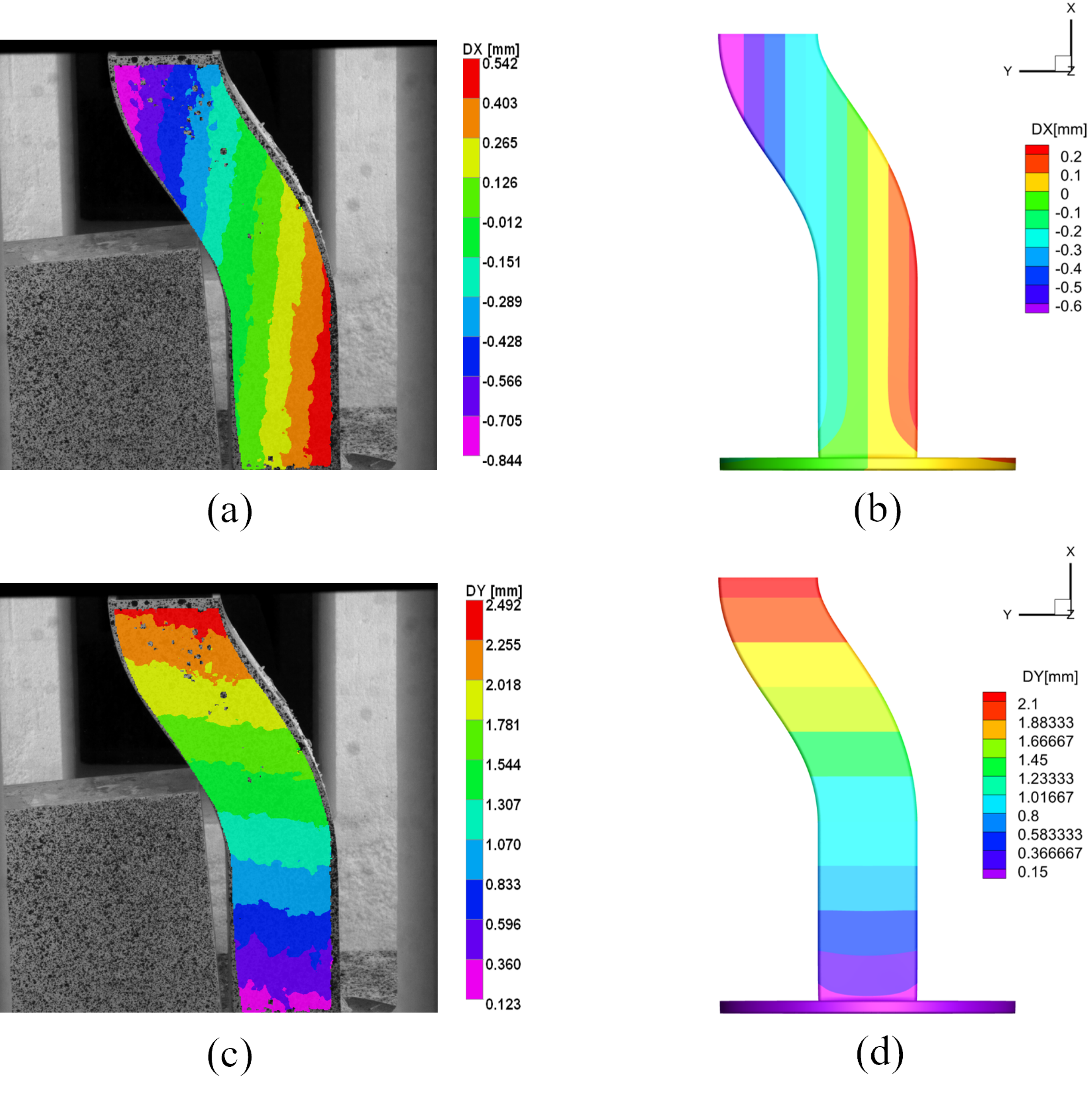}
\caption{Thermal deformation measurement results at 700℃. \textbf{(a)} Actual displacement U, \textbf{(b)} simulated displacement U, \textbf{(c)} actual displacement V, and \textbf{(d)} simulated displacement V}
 \label{fig10}
\end{figure}

To enable quantitative visual comparison between experimental measurements and numerical simulations, we standardized the colormap ranges and scaling for both datasets. As shown in Fig. \ref{fig9} and Fig. \ref{fig10}, the quantitative comparison between the experimental measurements and numerical simulations reveals consistent agreement in both deformation magnitude and evolutionary trends.  Experimental results demonstrate consistent deformation behavior at both 500°C and 700°C. Figures \ref{fig9}(a) and (b), figures \ref{fig10}(a) and (b) illustrate the horizontal displacement U, showing thermal expansion along the specimen centerline with negative values indicating expansion opposite to the defined coordinate system positive direction. Figures \ref{fig9}(c) and (d), and figures \ref{fig10}(c) and (d) present the vertical displacement V, revealing monotonic increase in upward thermal expansion magnitude with elevation from the fixed base, following a characteristic gradient pattern typical of thermally induced bending. The thermal deformation measurement results of the S-shaped specimen highlight the significant potential of the proposed method in engineering 3D deformation measurement.

\section{Conclusion}\label{sec5}

In this paper, we present a pose guidance method for stereo calibration, which guides users to complete high-precision stereo calibration through a graphical user interface. The method solves for the next optimal pose by introducing a joint optimization for relative and absolute extrinsic parameters and adopting the minimization of the trace of the covariance matrix as the loss function. The experimental results demonstrate that the proposed pose guidance method significantly enhances the precision and efficiency of stereo calibration in 3D deformation measurement. 

In experiments, this study systematically evaluates calibration efficiency and accuracy through comparative experiments on pose selection strategies. Simulation experiments have verified the robustness of this method. The calibration efficiency test demonstrates that our optimal pose method achieves comparable accuracy to random pose while requiring fewer calibration images and maintaining robustness across varying FOVs. In the accuracy validation of using a cylindrical specimen under 80-ton axial compression, the calibration of our method (2-random + 6-optimal) outperformed that of random pose (8-random), yielding lower reprojection error (0.0471 vs. 0.0531 pixels) and superior strain measurement precision (3.3 ± 18.2 $~\mu\varepsilon$ vs. 14.7 ± 30.2 $~\mu\varepsilon$ ). This dual improvement in both precision and efficiency validates the practical efficacy of the proposed optimization framework.

Finally, the proposed method, in conjunction with DIC, has been successfully applied to thermal deformation measurement tasks. By spraying a high-temperature-resistant speckle pattern on the surface of an S-shaped specimen and utilizing the stereo calibration method proposed in this paper, deformation measurement under high-temperature conditions was achieved. The measurement results show good agreement with the simulation results obtained from the software, further highlighting the significant potential in engineering deformation measurement.  The image distortion caused by heat haze is regarded as the primary source of error in our experiment. This limitation highlights a key challenge in the current experimental setup. Future work will focus on optimizing the system configuration and developing image processing techniques to enhance measurement stability under high-temperature conditions.

\bmhead{Acknowledgements}

This work is supported by Hunan Provincial Natural Science Foundation for Excellent Young Scholars (Grant 2023J120045), National Natural Science Foundation of China (Grant 12372189) and the Science and Technology Innovation Program of Hunan Province(2022RC1196). 

\section*{Declarations}
\textbf{Conflict of interest}   The authors declare that they have no conflict of interest.

\bibliography{sn-bibliography.bib}

\clearpage
\Large
\begin{center}
    {\bf Supplementary Material }
\end{center}
\normalsize
\setcounter{page}{1}

\begin{appendices}
\section{\label{secA}The explicit formulas of $\mathbf{U}_i$ and $\mathbf{V}_i$ }

In the following, we derive the analytical expression of the Jacobian blocks $\mathbf{U}_i$ and $\mathbf{V}_i$. Initially, we compute the Jacobian block ($\mathbf{V}_i$) of the left camera's reprojection error with respect to its absolute extrinsic parameters, which characterizes the influence of pose perturbations on projective geometry. 

Assuming that the 3D coordinates of the control point $\mathbf{P}({X^w},{Y^w},{Z^w})$ in the left camera coordinate system are 
${\mathbf{Q}^l}({X^l},{Y^l},{Z^l})$, the rigid-body transformation between the left camera coordinate system and the world coordinate system is given by:
\begin{equation}
{\mathbf{Q}^l} = {\mathbf{R}^l}\mathbf{P} + {\mathbf{t}^l}
\end{equation}

\noindent where ${\mathbf{R}^l}$ denotes the rotation matrix, and the rotation vector in the Lie algebra \cite{bib38}. We employ the following model to characterize lens distortion effects:
\begin{equation}
\left\{ {\begin{array}{*{20}{c}}
{x_d^l = {x^l}(1 + k_1^l{{({r^l})}^2} + k_2^l{{({r^l})}^4}) + 2p_1^l{x^l}{y^l} + p_2^l({{({r^l})}^2} + 2{{({x^l})}^2})}\\
{y_d^l = {y^l}(1 + k_1^l{{({r^l})}^2} + k_2^l{{({r^l})}^4}) + p_1^l({{({r^l})}^2} + 2{{({x^l})}^2}) + 2p_2^l{x^l}{y^l}}
\end{array}} \right.
\end{equation}

\noindent with 
\begin{equation}
{x^l} = \frac{{{X^l}}}{{{Z^l}}}, \quad {y^l} = \frac{{{Y^l}}}{{{Z^l}}}, \quad {({r^l})^2} = {({x^l})^2} + {({y^l})^2}
\end{equation}

For clarity of presentation, we define ${\mathbf{x}^l}({x^l},{y^l})$ as the normalized coordinate and $\mathbf{x}_d^l(x_d^l,y_d^l)$ as the distorted normalized coordinate. Based on the projection model, $\mathbf{x}_d^l$ can be mapped to the pixel coordinates ${\mathbf{p}^l}({u^l},{v^l})$:
\begin{equation}
\left\{ {\begin{array}{*{20}{c}}
{{u^l} = f_x^lx_d^l + c_x^l}\\
{{v^l} = f_y^ly_d^l + c_y^l}
\end{array}} \right.
\end{equation}

\noindent where $f_x^l$, $f_y^l$, $c_x^l$ and $c_y^l$ are the intrinsic parameters of the left camera. As described in this paper, the reprojection residual ${\mathord{\buildrel{\lower3pt\hbox{$\scriptscriptstyle\frown$}}\over {\mathbf{p}}} ^l}(p_x^l,p_y^l)$  characterizes the differentials between the observed coordinates ${\tilde {\mathbf{p}}^l}({\tilde u^l},{\tilde v^l})$ and the projected coordinates ${{\mathbf{p}}^l}({u^l},{v^l})$:
\begin{equation}
\left\{ {\begin{array}{*{20}{c}}
{p_x^l = {{\tilde u}^l} - {u^l}}\\
{p_y^l = {{\tilde v}^l} - {v^l}}
\end{array}} \right.
\end{equation}

Our goal is to derive the partial derivative of the left camera's reprojection residual with respect to its absolute extrinsic parameters. According to the chain rule, it can be expressed as:
\begin{equation}
\frac{{\partial {{\mathord{\buildrel{\lower3pt\hbox{$\scriptscriptstyle\frown$}} 
\over {\mathbf{p}}} }^l}}}{{\partial {{\mathbf{R}}^l}}} = \frac{{\partial {{\mathord{\buildrel{\lower3pt\hbox{$\scriptscriptstyle\frown$}} 
\over {\mathbf{p}}} }^l}}}{{\partial {\mathbf{x}}_d^l}} \cdot \frac{{\partial {\mathbf{x}}_d^l}}{{\partial {{\mathbf{x}}^l}}} \cdot \frac{{\partial {{\mathbf{x}}^l}}}{{\partial {{\mathbf{Q}}^l}}} \cdot \frac{{\partial {{\mathbf{Q}}^l}}}{{\partial {{\mathbf{R}}^l}}}
\end{equation}

\noindent and
\begin{equation}
\frac{{\partial {{\mathord{\buildrel{\lower3pt\hbox{$\scriptscriptstyle\frown$}} 
\over {\mathbf{p}}} }^l}}}{{\partial {{\mathbf{t}}^l}}} = \frac{{\partial {{\mathord{\buildrel{\lower3pt\hbox{$\scriptscriptstyle\frown$}} 
\over {\mathbf{p}}} }^l}}}{{\partial {\mathbf{x}}_d^l}} \cdot \frac{{\partial {\mathbf{x}}_d^l}}{{\partial {{\mathbf{x}}^l}}} \cdot \frac{{\partial {{\mathbf{x}}^l}}}{{\partial {{\mathbf{Q}}^l}}} \cdot \frac{{\partial {{\mathbf{Q}}^l}}}{{\partial {{\mathbf{t}}^l}}}
\end{equation}

The solutions for each component are derived as follows:
\begin{equation}
\frac{{\partial {{\mathbf{p}}^l}}}{{\partial {\mathbf{x}}_d^l}} =  - \left[ {\begin{array}{*{20}{c}}
{f_x^l}&0\\
0&{f_y^l}
\end{array}} \right]
\end{equation}

\begin{equation}
\frac{{\partial {\mathbf{x}}_d^l}}{{\partial {{\mathbf{x}}^l}}} = \left[ {\begin{array}{*{20}{c}}
{\frac{{\partial x_d^l}}{{\partial {x^l}}}}&{\frac{{\partial x_d^l}}{{\partial {y^l}}}}\\
{\frac{{\partial y_d^l}}{{\partial {x^l}}}}&{\frac{{\partial y_d^l}}{{\partial {y^l}}}}
\end{array}} \right]
\end{equation}

\noindent with
\begin{equation}
\left\{ {\begin{array}{*{20}{c}}
{\frac{{\partial x_d^l}}{{\partial {x^l}}} = (1 + k_1^l{{({r^l})}^2} + k_2^l{{({r^l})}^4}) + 2{{({x^l})}^2}(k_1^l + 2k_2^l{{({r^l})}^2}) + 2p_1^l{y^l} + 6p_2^l{x^l}}\\
{\frac{{\partial y_d^l}}{{\partial {y^l}}} = (1 + k_1^l{{({r^l})}^2} + k_2^l{{({r^l})}^4}) + 2{{({y^l})}^2}(k_1^l + 2k_2^l{{({r^l})}^2}) + 6p_1^l{y^l} + 2p_2^l{x^l}}\\
{\frac{{\partial x_d^l}}{{\partial {y^l}}} = 2{x^l}{y^l}(k_1^l + 2k_2^l{{({r^l})}^2}) + 2p_1^l{x^l} + 2p_2^l{y^l}}\\
{\frac{{\partial y_d^l}}{{\partial {x^l}}} = 2{x^l}{y^l}(k_1^l + 2k_2^l{{({r^l})}^2}) + 2p_1^l{x^l} + 2p_2^l{y^l}}
\end{array}} \right.
\end{equation}

\begin{equation}
\frac{{\partial {{\mathbf{x}}^l}}}{{\partial {{\mathbf{Q}}^l}}} = \left[ {\begin{array}{*{20}{c}}
{\frac{1}{{{Z^l}}}}&0&{ - \frac{{{X^l}}}{{{{({Z^l})}^2}}}}\\
0&{\frac{1}{{{Z^l}}}}&{ - \frac{{{Y^l}}}{{{{({Z^l})}^2}}}}
\end{array}} \right]
\end{equation}

\begin{equation}
\frac{{\partial {{\mathbf{Q}}^l}}}{{\partial {{\mathbf{R}}^l}}} =  - {\left[ {{{\mathbf{Q}}^l}} \right]_ \times }, \quad 
\frac{{\partial {{\mathbf{Q}}^l}}}{{\partial {{\mathbf{t}}^l}}} = {\mathbf{I}}
\end{equation}

Thus, we have rigorously derived the solution for the Jacobian block ($\mathbf{V}_i$) of the left camera's reprojection residual with respect to its absolute extrinsic parameters. While the imaging geometry of the right camera is analogous to the left camera, its absolute extrinsic parameters are represented as a composite transformation of the left camera's absolute extrinsic parameters and their relative extrinsic parameters. Due to the high-precision estimation of the relative extrinsic parameters constitutes a core objective of stereo calibration, we therefore define the Jacobian block $\mathbf{U}_i$ as the partial derivative of the right camera's reprojection residual with respect to the relative extrinsic parameters. 

In fact, the calculation of $\mathbf{U}_i$ follows a similar procedure, with the only difference being that the left camera coordinate system is used as the world coordinate system, that is:
\begin{equation}
\mathbf{Q}^{r} = \mathbf{R}\mathbf{Q}^{l} + \mathbf{t} = \mathbf{R}(\mathbf{R}^{l}\mathbf{P} + \mathbf{t}^{l}) + \mathbf{t}
\end{equation}

\noindent the solutions for the relevant components are modified accordingly:
\begin{equation}
\frac{\partial \mathbf{Q}^r}{\partial \mathbf{R}} = -{\left[ \mathbf{Q}^r \right]}_\times, \quad \frac{\partial \mathbf{Q}^r}{\partial \mathbf{t}} = \mathbf{I}
\end{equation}

All the parameters of the Jacobian matrix are now determined.

\section{\label{secB}Covariance derivation}

Let the joint covariance $\begin{pmatrix}
\bm{\theta}_{\mathrm{rel}} \\
\bm{\theta}_{\mathrm{abs}}
\end{pmatrix}$ be defined by the equation matrix:
\begin{equation}
\mathrm{Cov}\begin{pmatrix}
\bm{\theta}_{\mathrm{rel}} \\ 
\bm{\theta}_{\mathrm{abs}}
\end{pmatrix}
= \left( \mathbf{J}^\mathsf{T} \mathbf{J} \right)^{-1}
\end{equation}

\noindent where $\bm{\theta}_{\mathrm{rel}} = \begin{bmatrix}
\mathbf{R}, & \mathbf{t}
\end{bmatrix}$ and $\bm{\theta}_{\mathrm{abs}} = \begin{bmatrix}
\mathbf{R}_i^l, & \mathbf{t}_i^l
\end{bmatrix}$ are positive definite.

According to Eq. \eqref{equ7} in our manuscript, the information matrix $\mathbf{J}^\mathsf{T} \mathbf{J}$ can be determined as:
\begin{equation}
\mathbf{J}^\mathsf{T}\mathbf{J} = 
\begin{pmatrix}
\sum\limits_{i=1}^m \mathbf{U}_i^\mathsf{T}\mathbf{U}_i & \mathbf{U}_1^\mathsf{T}\mathbf{V}_1 & \mathbf{U}_2^\mathsf{T}\mathbf{V}_2 & \cdots & \mathbf{U}_m^\mathsf{T}\mathbf{V}_m \\
\mathbf{V}_1^\mathsf{T}\mathbf{U}_1 & \mathbf{V}_1^\mathsf{T}\mathbf{V}_1 & 0 & \cdots & 0 \\
\mathbf{V}_2^\mathsf{T}\mathbf{U}_2 & 0 & \mathbf{V}_2^\mathsf{T}\mathbf{V}_2 & \cdots & 0 \\
\vdots & \vdots & \vdots & \ddots & \vdots \\
\mathbf{V}_m^\mathsf{T}\mathbf{U}_m & 0 & 0 & \cdots & \mathbf{V}_m^\mathsf{T}\mathbf{V}_m
\end{pmatrix}
\end{equation}

For clarity of presentation, we construct the block matrices as given in Eqs \eqref{equ8} and \eqref{equ9} of the manuscript. According to the Schur Complement Lemma \cite{bib39}, for a block matrix,
\begin{equation}
\mathbf{M} = \begin{pmatrix}
\mathbf{E} & \mathbf{F} \\
\mathbf{G} & \mathbf{H}
\end{pmatrix}
\end{equation}

\noindent if $\mathbf{H}$ is invertible, then the Schur complement of $\mathbf{E}$  with respect to $\mathbf{M}$  is well-defined and given by:
\begin{equation}
\bm{\Sigma} = (\mathbf{E} - \mathbf{F}\mathbf{H}^{-1}\mathbf{G})^{-1}
\end{equation}

For our case, the block matrix is defined as:
\begin{equation}
\mathbf{J}^\mathsf{T}\mathbf{J} = 
\begin{pmatrix}
\mathbf{A} & \mathbf{C} \\
\mathbf{C}^\mathsf{T} & \mathbf{B}
\end{pmatrix}
\end{equation}

\noindent since $\mathbf{B}$  is block-diagonal and each block $\mathbf{V}_i^\mathsf{T} \mathbf{V}_i$ is invertible, due to the positive definiteness of measurement Jacobians, $\mathbf{B}$ is invertible. By the definition of the joint covariance, the sub-block corresponding to $\bm{\theta}_{\mathrm{rel}}$ can be described as:
\begin{equation}
\mathrm{Cov}(\bm{\theta}_{\mathrm{rel}}) = (\mathbf{A} - \mathbf{C}\mathbf{B}^{-1}\mathbf{C}^\mathsf{T})^{-1} = \bm{\Sigma}
\end{equation}

\section{\label{secC}Improvement over independent monocular calibration}
To quantify how stereo-specific constraints improve performance over independent monocular calibration, we have conducted relevant comparative experiments for verification. And all calibration images are checkerboard images captured under optimal pose guidance, and the intrinsic parameters and distortion coefficients are kept consistent. The monocular calibration procedure is defined as: solving the absolute extrinsic parameters of the left and right cameras separately from each corresponding set of calibration images, and then deriving relative extrinsic parameters from the absolute extrinsic parameters of the left and right cameras based on the variant of Eq. \eqref{equ3}. Perform a quantitative comparison between the mean value of derived relative extrinsic parameters and the relative extrinsic parameters obtained from our method. Note that the translation vector is calculated using the arithmetic mean method, whereas the rotation matrices need to first be converted into quaternions, averaged in quaternion space, and then transformed back into rotation vectors. The calibration images used are the 2-random + 6-optimal images in Section \ref{subsec42}, and the result is shown as follows:

\begin{table}[htbp]
  \centering
  \caption{Quantitative comparison of relative extrinsic parameters between independent monocular calibration and stereo calibration}
  \begin{tabular}{ccccc}
    \toprule
    \textbf{Calibration} & \multirow{2}{*}{$\boldsymbol{\mathbf{R} / \text{deg}}$} & \multirow{2}{*}{$\boldsymbol{\mathbf{t} / \text{mm}}$} & \multirow{2}{*}{$\boldsymbol{\mathbf{RE} / \text{pixel}}$} & \multirow{2}{*}{$\boldsymbol{\mathbf{TE} / \text{mm}}$} \\
    \textbf{scheme} &  &  &  &  \\
    \midrule
    Monocular & $(1.181, -19.421, -1.104)$ & $(-370.4, -12.7, 31.7)$ & $0.2502$ & $0.1703$ \\
    \midrule
    Stereo & $(-0.388, -19.186, -0.532)$ & $(-385.0, -5.8, 72.0)$ & $0.2715$ & $0.0676$ \\
    \bottomrule
  \end{tabular}
  \label{tab_1}
\end{table}

As shown in Table \ref{tab_1}, according to the above procedure, although the reprojection error of stereo calibration realized from monocular calibration results is smaller than that of our method, the triangulation error calculated therefrom is much larger. We consider that there is a certain overfitting phenomenon in its calibration results. Compared with stereo calibration, monocular calibration lacks some geometric constraints between cameras, so it is more prone to overfitting.

\section{\label{secD}Compare against stereo-specialized methods}

\begin{table}[htbp]
\centering
\caption{Comparison of different stereo calibration methods}
\label{tab_2}
\begin{tabular*}{\textwidth}{@{\extracolsep{\fill}}c@{\extracolsep{\fill}}c@{\extracolsep{\fill}}c@{}}
\hline
\textbf{Parameter} & \textbf{Our} & \textbf{Guan[10]} \\
\hline
$\boldsymbol{\mathbf{R}}$ & $(-0.388, -19.186, -0.532)$ & $(-0.407, -19.271, -0.509)$ \\
\hline
$\boldsymbol{\mathbf{t}}$ & $(-385.0, -5.8, 72.0)$ & $(-390.2, -10.1, 65.3)$ \\
\hline
$\text{RE} / \text{pixel}$ & $0.2715$ & $0.2924$ \\
\hline
$\text{TE} / \text{mm}$ & $0.0676$ & $0.0734$ \\
\hline
\end{tabular*}
\end{table}

We have compared our method with the stereo calibration method in \cite{bib17}. This method estimates intrinsic and extrinsic parameters simultaneously using nine or more image points based on the epipolar constraint between stereo image pairs. Since it is a self-calibration method, it does not require the use of a checkerboard. To enable an effective comparison between the two methods, we have used the intersection coordinates extracted from the checkerboard as input to the method in  \cite{bib17}. The calibration images used are also the 2-random + 6-optimal images in Sec. \ref{subsec42}:

As shown in Table \ref{tab_2}, compared with the method in  \cite{bib17}, our method has slight advantages in both reprojection error and intersection error, which demonstrates the effectiveness of our method. Meanwhile, both stereo calibration methods outperform independent monocular calibration in terms of triangulation error, which also reflects the superiority of the binocular system in measurement.

\end{appendices}

\end{document}